\documentclass{article}
\usepackage{tabularx}
\usepackage{array}
\usepackage{arxiv}
\usepackage{graphicx}
\usepackage[utf8]{inputenc} 
\usepackage[T1]{fontenc}    
\usepackage{hyperref}       
\usepackage{url}            
\usepackage{booktabs}       
\usepackage{amsfonts}       
\usepackage{nicefrac}       
\usepackage{microtype}      
\usepackage{lipsum}
\usepackage{natbib}
\usepackage{verbatim}
\title{GenNet: Reading Comprehension with Multiple Choice Questions using Generation and Selection model}

\author{
  Vaishali Ingale\\
  Department of Information Technology\\
  Army Institute of Technology, Pune\\
 \texttt{vingale@aitpune.edu.in} \\
   \And
Pushpender Singh\\
 Department of Information Technology\\
  Army Institute of Technology, Pune\\
  \texttt{pushpender\_16372@aitpune.edu.in} \\
}

\begin{document}
\maketitle

\begin{abstract}
Multiple-choice machine reading comprehension is difficult task as its required machines to select the correct option from a set of candidate or possible options using the given passage and question.Reading Comprehension with Multiple Choice Questions task, required a human (or machine) to read a given passage, question pair and select the best one option from n given options. There are two different ways to select the correct answer from the given passage. Either by selecting the best match answer to by eliminating the worst match answer. Here we proposed GenNet model, a neural network-based model. In this model first we will generate the answer of the question from the passage and then will matched the generated answer with given answer, the best matched option will be our answer. For answer generation we used S-net \citep{tan2017s} model trained on SQuAD and To evaluate our model we used Large-scale  RACE (ReAding Comprehension Dataset From Examinations) \citep{lai2017race}.
\end{abstract}

\section{Introduction}
\begin{figure}[ht]
\centering
\includegraphics[width=10 cm]{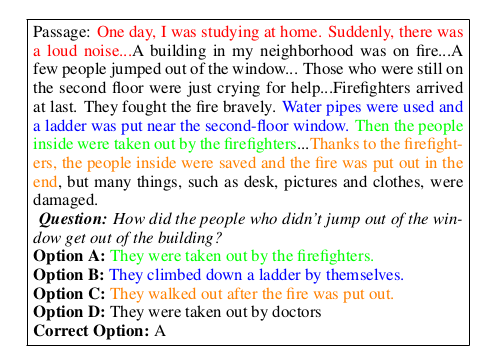}
\caption{An example multiple-choice reading comprehension question. }
\label{fig1}
\end{figure}

Reading comprehension is one of the fundamental skills for human, which one learn systematically since the elementary school. Reading comprehension give human the ability of reading texts, understanding their meanings,and with the help of given context answering questions. When
machines are required to comprehend texts, they first need to
understand the unstructured text and do reasoning based on given
text \citep{chen2016thorough}\citep{wang2018using}.Answering questions based a passage requires an individual unique skill set. It requires ability to perform basic mathematical operations and logical ability (e.g. to answer questions
like how many times Amit visited sweet shop?), look-up ability, ability to deduce, ability to gather information contained in multiple sentences and passages. This diverse and unique skill set makes question answering a challenging task.There are several variants of this task, For example, if we have a given passage and a question, the answer could either (i) be generated from the passage (ii) match some span in the passage (iii) or could be one of the n number of given candidate answers. The last variant is mostly used in various high school, quiz , middle school, and different competitive examinations. This variant of Reading Comprehension generally referred as Reading Comprehension with Multiple Choice Questions (RC-MCQ).In the given figure \ref{fig1} We have a passage and a question and 4 candidate answers. Task here defined is to find the most suitable answer from the passage for given question. While answering such Multiple Choice Questions (MCQs) figure \ref{fig1}, humans typically use a combination of option elimination and option selection or sometimes they find answer from the passage i.e they generate the answer of the question from passage and match the generated answer with given options and they choose more close candidate as correct answer.
\par
\begin{figure}[ht]
\centering
\includegraphics[width=10 cm]{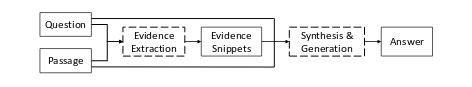}
\caption{Overview of S-Net.\citep{tan2017s}}
\label{fig2}
\end{figure}
\begin{figure}[ht]
\centering
\includegraphics[width=10 cm]{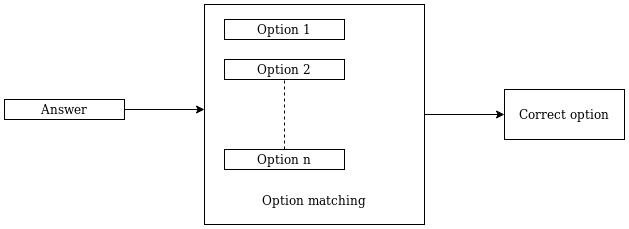}
\caption{Overview of option matching and selection. }
\label{fig3}
\end{figure}

Here we proposed model which mimic the answer generation and then matching human process.First the span where possible answer in the passage is computed. we first compute a question-aware representation of the passage (which essentially tries to retain portions of the passage which are only relevant to the question). Then we use answer generation using state-of-art S-Net model \citep{tan2017s}which extract and generate answer figure \ref{fig2}. After we have answer generated from the passage now we  weight every given candidate option and select the best matched option. That best matched option was our answer figure \ref{fig3}.

\section{Related Work}
Datasets played an important role in machine reading comprehension, there were different type of datasets designed to solve different variant of machine reading comprehension.  SQuAD dataset\citep{rajpurkar2016squad} was designed to answer simple question answer reading comprehension that aims to answer a question with exact text spans in a passage. Later MS-MACRO dataset\citep{nguyen2016ms} was designed for multi-passage reading comprehension. CNN/ Dailymail \citep{chen2016thorough} and Who did what dataset\citep{onishi2016did} designed for cloze variant reading comprehension. MCtest\citep{richardson2013mctest} and RACE dataset\citep{lai2017race} are released for Multiple choice question variant  reading comprehension.
\par
Similar work in reading comprehension where Multiple choice variant of Comprehension considered includes Hierarchical Attention Flow model\citep{zhu2018hierarchical}, in this model the candidate options leverage  to model the interaction between question options and passage.This was a option selection model which select the correct option from the given candidate options. Other work relatable to this paper was eliminating options model\citep{parikh2019eliminet} which eliminate the wrong answer from the candidate answer.Multi matching network\citep{tang2019multi} models interaction relationship between passage, questions and candidate answer. It take different paradigm of matching into account. Option comparison Network \citep{ran2019option} compares between options at word level and identify correlation to help buildup logic and reasoning. 
Co-matching model \citep{wang2018co} is used to match between answer and question and passage pair. It match for the relationship between question and answer with the passage. Bidirectional co-matching based model \citep{zhang2019dual} matched passage and question, answer bidirectionally. The Convolutional Spatial Attention (CSA) model \citep{chen2019convolutional} form the enriched representaion by  fully extract the mutual information among the passage, question, and the candidates.
\par
To generate answer several models are there like QANet \citep{yu2018qanet} which combined local Convolution with Global Self-Attention and its encoder consist exclusively of convolution and self-attention.Bidirectional Attention Flow model \citep{seo2016bidirectional} use to focus on large span of passage. BIDAF network is a multi stage hierarchical process and use bidirection attention flow to obtain a query-aware context representation. But the reason to use S-Net model as answer generation model because S-Net not only find the answer from the passage but it can also synthesise passage when required. Some questions are tricky and there answer lies in different span of passage. In such situation S-Net is useful as it remember the past context for longer time as it have GRU as basic component.

\section{Proposed model}
\begin{figure}[ht]
\centering
\includegraphics[width=10 cm]{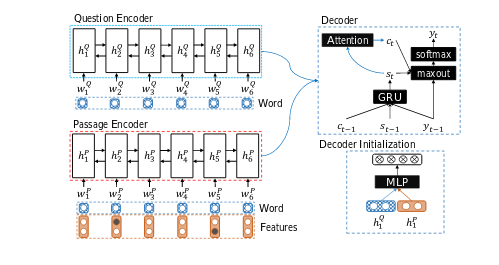}
\caption{Answer Synthesis/Generation Model\citep{tan2017s} }
\label{fig4}
\end{figure}

There are two tasks needs to be performed in this model. First is Answer extraction and Answer Synthesis/Generation and then option selection.
Answer extraction and Generation will be done using state-of-art S-NET model\citep{tan2017s}.  S-Net first pull out evidence snippets by matching the question and passage respectively, and then generates the answer by filtering the question, passage, and evidence snippets. consider a passage \(P =[p_1,p_2,p_3,...p_p]\) of word length P, Question \(Q =[Q_1,Q_2,Q_3,...Q_q]\) of word length Q, and n options \(Z_n =[z_1,z_2,z_3,...z_k]\) where n > 1 and word length k.
We first convert the words to their word-level embedding and character-level embedding using GLOVE\citep{pennington2014glove}.The encoding and embedding layers take in a series of tokens and represent it as a series of vectors. The character-level embeddings are cause by taking the final hidden states of a bi-directional GRU applied to embedding of characters in the token. They then use a bi-directional Gated Recurrent Unit to give rise to new depiction \(u_1^p,u_2^p,u_3^p,...u_p^p\) for questions as well as  \(u_1^q,u_2^q,u_3^q,...u_q^q\) for passages too and \(u_1^z,u_2^z,u_3^z,...u_z^z\) for options as well. The embedding matrix is boot only once and not trained in the entire learning process. As shown in Figure \ref{fig4} S-NET uses the series-to-series model to incorporate the answer with the extracted evidences as features.  They first produce the depiction
It first produce the depiction \(h_p^t\) and \(h_q^t\) of all words in the question and passage respectively. When giving out the answer depiction, it merge the basic word embedding \(e_p^t\) with some added features \(f_s^t\) and \(f_e^t\)
to indicate the end and start place of the
evidence snippet given out by evidence extraction model. \(f_s^t\)=1 and \(f_e^t\)=1 mean the position
t is the start and end of the evidence span, respectively.
\begin{equation}
    h_t^p=BiGRU(h_{t-1}^p,[e_t^p,f_t^s,f_t^e])
\end{equation}
    \begin{equation}
        h_t^q=BiGRU(h_{t-1}^q,e_t^q)
    \end{equation}

On top of the encoder, S-Net uses GRU with attention as the decoder to produce the answer.  At each decoding time step t , the GRU reads the previous word embedding \(w_{t-1}\) and previous context vector \(c_{t-1}\) and finally produced answer.

The produced answer will be stored in Answer vector.
 \(A_n =[a_1,a_2,a_3,...a_a]\) where a is length of the answer.Figure \ref{fig3} shows the overview of selection module. The selection module will take the refined  answer  representation
\(a_t\) and computes its  bi-linear similarity with each option representation.

\begin{equation}
    score(i)=a_tW_{att}z_{t_i}
\end{equation}
where i is the number of option, \(a_t\) is generated answer vector, \(z_{t_i}\) is option vector and \(W_{att}\)
is  a  matrix  which needs  to  be  learned.   We  select  the  option  which  gives  the highest score as computed above.  We train the model using the cross entropy loss by normalizing the above scores (using softmax) first to obtain a probability distribution.

\section{Experimental Setup}
Here we discussed about the dataset used to evaluate our model, Training procedure, result comparison and future work.
\subsection{Dataset}
 We evaluate our model on RACE  dataset\citep{lai2017race}
 Race is a large-scale reading comprehension dataset with more than 28,000 passages and nearly 100,000 questions. The dataset is collected from English examinations in China, which are designed for middle school and high school students. Each passage is a JSON file. The JSON file contains fields (i) article: A string, which is the passage (ii) questions: A string list. Each string is a query. There are two types of questions. First one is an interrogative sentence. Another one has a placeholder, which is represented by \_. (iii)options: A list of the options list. Each options list contains 4 strings, which are the candidate option. (iv) answers: A list contains the golden label of each query.(v) id: Each passage has a unique id in this dataset. RACE has wide variety of questions like Summarization, Inference, Deduction and Context matching.

\begin{figure}[ht]
\centering
\includegraphics[width=10 cm]{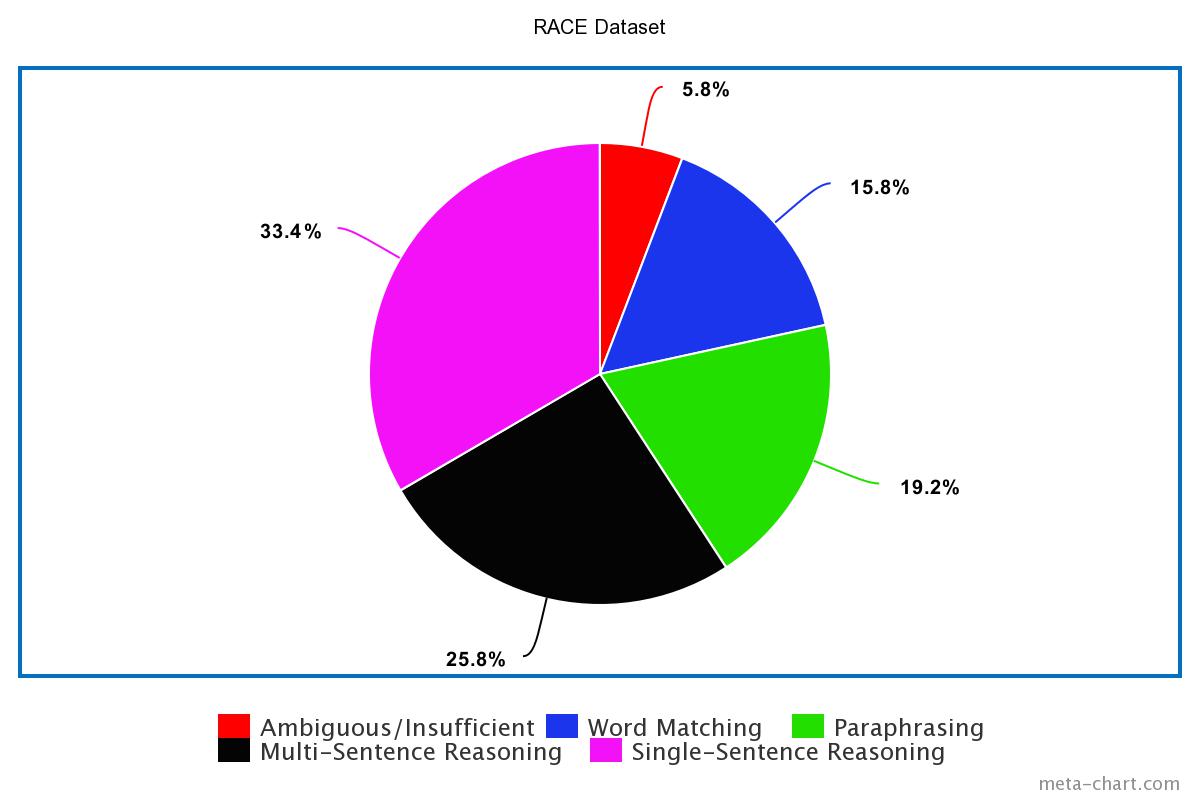}
\caption{Statistic information about Reasoning type in RACE dataset }
\label{fig5}
\end{figure}

\subsection{Training Procedures and Hyper-parameter}
We integrate two different model into once. First we train our model on S-Net. To train model on S-Net we process dataset differently. We only consider passage and question and correct option to train model on S-Net. Later we pass the result on to next stage on our model where we train model using generated answer and all candidate options.
To train the model, we used stochastic gradient descent with ADAM
optimizer.\citep{kingma2014adam} We initialize learning rate
with 0.005. Gradients are clipped in L2-norm to no larger than
10. To update model parameter per step,we used A mini-batch of 32 samples. We have created a vocabulary using top 65k words from passage and questions and if a new out-of-vocabulary(OOV) word encountered we add a special token UNK. We use the same vocabulary for the passage, question, and options vector embedding. We tune all our models based  on  the  accuracy  achieved  on  the  validation  set. We use 300 dimensional Glove embedding \citep{pennington2014glove} for word embedding and word and character encoding.We experiment with
both fine-tuning and not fine-tuning these word embedding. We train all our models for upto 80 epochs as
we do not see any benefit of training beyond 80 epochs as result were starting recurrence.The hidden state size
of all GRU network is 128. We apply dropout \citep{srivastava2014dropout}to word embeddings and BiGRU’s outputs with a drop rate of 0.45.

\subsection{Results and Future Work}
\begin{table}[h]
    \centering
    \begin{tabular}{l c l c l c l c}
    \hline
    \toprule
         Model & RACE-Mid & RACE-High & RACE \\
         \midrule
         Random* & 24.6& 25.0 & 24.9\\
         Sliding Window*& 37.3 & 30.4 & 32.2\\
         GA Reader (100D)* & 43.7 & 44.2 & 44.1\\
        Stanford AR (100D)* & 44.2 & 43.0 & 43.3\\
        Sliding Window* & 37.3 & 30.4 & 32.2\\
         \textbf{GenNet} & \textbf{79.6} & \textbf{75.4} &\textbf{77.3}\\
    \bottomrule
    \\
    \end{tabular}
    \caption{Accuracy on test set of RACE-M, RACE-H and
RACE. * indicates the results from \citep{lai2017race} which are trained with 100D pre-trained Glove word embeddings}
    \label{tab:my_label}
\end{table}
The Human Ceiling Performance reported by CMU on RACE dataset is 
94.2. Our model gives accuracy of 79.6 \% on RACE-M 75.4 \% on RACE-H and 77.3\% on RACE FULL which outperform several other model. 
Since in this model first answer are generated and then option is selected such model can be used to solve such multiple choice question whose answer option is not present or MCQ with "none of the above" or "No answer" type multiple choice questions. 
\section{Conclusion}
In this paper, we present the GenNet model for
multiple-choice reading comprehension. Specifically, the model uses
a  combination  of  Generation  and  selection  to  arrive  at  the
correct option. This is achieved by first generating the answer for the questions from the passage and then matching generated answer with the options.At last, the proposed model achieves overall sate-of-the-art accuracy on RACE and significantly outperforms  neural network baselines on  RACE-M, RACE-H and RACE FULL.As future work, we would like to work towards unanswerable questions or questions where no option matched.
\bibliographystyle{apalike}
\bibliography{references}

\begin{thebibliography}{}

\bibitem[Chen et~al., 2016]{chen2016thorough}
Chen, D., Bolton, J., and Manning, C.~D. (2016).
\newblock A thorough examination of the cnn/daily mail reading comprehension
  task.
\newblock {\em arXiv preprint arXiv:1606.02858}.

\bibitem[Chen et~al., 2019]{chen2019convolutional}
Chen, Z., Cui, Y., Ma, W., Wang, S., and Hu, G. (2019).
\newblock Convolutional spatial attention model for reading comprehension with
  multiple-choice questions.
\newblock In {\em Proceedings of the AAAI Conference on Artificial
  Intelligence}, volume~33, pages 6276--6283.

\bibitem[Kingma and Ba, 2014]{kingma2014adam}
Kingma, D.~P. and Ba, J. (2014).
\newblock Adam: A method for stochastic optimization.
\newblock {\em arXiv preprint arXiv:1412.6980}.

\bibitem[Lai et~al., 2017]{lai2017race}
Lai, G., Xie, Q., Liu, H., Yang, Y., and Hovy, E. (2017).
\newblock Race: Large-scale reading comprehension dataset from examinations.
\newblock {\em arXiv preprint arXiv:1704.04683}.

\bibitem[Nguyen et~al., 2016]{nguyen2016ms}
Nguyen, T., Rosenberg, M., Song, X., Gao, J., Tiwary, S., Majumder, R., and
  Deng, L. (2016).
\newblock Ms marco: a human-generated machine reading comprehension dataset.

\bibitem[Onishi et~al., 2016]{onishi2016did}
Onishi, T., Wang, H., Bansal, M., Gimpel, K., and McAllester, D. (2016).
\newblock Who did what: A large-scale person-centered cloze dataset.
\newblock {\em arXiv preprint arXiv:1608.05457}.

\bibitem[Parikh et~al., 2019]{parikh2019eliminet}
Parikh, S., Sai, A.~B., Nema, P., and Khapra, M.~M. (2019).
\newblock Eliminet: A model for eliminating options for reading comprehension
  with multiple choice questions.
\newblock {\em arXiv preprint arXiv:1904.02651}.

\bibitem[Pennington et~al., 2014]{pennington2014glove}
Pennington, J., Socher, R., and Manning, C.~D. (2014).
\newblock Glove: Global vectors for word representation.
\newblock In {\em Proceedings of the 2014 conference on empirical methods in
  natural language processing (EMNLP)}, pages 1532--1543.

\bibitem[Rajpurkar et~al., 2016]{rajpurkar2016squad}
Rajpurkar, P., Zhang, J., Lopyrev, K., and Liang, P. (2016).
\newblock Squad: 100,000+ questions for machine comprehension of text.
\newblock {\em arXiv preprint arXiv:1606.05250}.

\bibitem[Ran et~al., 2019]{ran2019option}
Ran, Q., Li, P., Hu, W., and Zhou, J. (2019).
\newblock Option comparison network for multiple-choice reading comprehension.
\newblock {\em arXiv preprint arXiv:1903.03033}.

\bibitem[Richardson et~al., 2013]{richardson2013mctest}
Richardson, M., Burges, C.~J., and Renshaw, E. (2013).
\newblock Mctest: A challenge dataset for the open-domain machine comprehension
  of text.
\newblock In {\em Proceedings of the 2013 Conference on Empirical Methods in
  Natural Language Processing}, pages 193--203.

\bibitem[Seo et~al., 2016]{seo2016bidirectional}
Seo, M., Kembhavi, A., Farhadi, A., and Hajishirzi, H. (2016).
\newblock Bidirectional attention flow for machine comprehension.
\newblock {\em arXiv preprint arXiv:1611.01603}.

\bibitem[Srivastava et~al., 2014]{srivastava2014dropout}
Srivastava, N., Hinton, G., Krizhevsky, A., Sutskever, I., and Salakhutdinov,
  R. (2014).
\newblock Dropout: a simple way to prevent neural networks from overfitting.
\newblock {\em The journal of machine learning research}, 15(1):1929--1958.

\bibitem[Tan et~al., 2017]{tan2017s}
Tan, C., Wei, F., Yang, N., Du, B., Lv, W., and Zhou, M. (2017).
\newblock S-net: From answer extraction to answer generation for machine
  reading comprehension.
\newblock {\em arXiv preprint arXiv:1706.04815}.

\bibitem[Tang et~al., 2019]{tang2019multi}
Tang, M., Cai, J., and Zhuo, H.~H. (2019).
\newblock Multi-matching network for multiple choice reading comprehension.
\newblock In {\em Proceedings of the AAAI Conference on Artificial
  Intelligence}, volume~33, pages 7088--7095.

\bibitem[Wang et~al., 2018a]{wang2018co}
Wang, S., Yu, M., Chang, S., and Jiang, J. (2018a).
\newblock A co-matching model for multi-choice reading comprehension.
\newblock {\em arXiv preprint arXiv:1806.04068}.

\bibitem[Wang et~al., 2018b]{wang2018using}
Wang, Y., Li, R., Zhang, H., Tan, H., and Chai, Q. (2018b).
\newblock Using sentence-level neural network models for multiple-choice
  reading comprehension tasks.
\newblock {\em Wireless Communications and Mobile Computing}, 2018.

\bibitem[Yu et~al., 2018]{yu2018qanet}
Yu, A.~W., Dohan, D., Luong, M.-T., Zhao, R., Chen, K., Norouzi, M., and Le,
  Q.~V. (2018).
\newblock Qanet: Combining local convolution with global self-attention for
  reading comprehension.
\newblock {\em arXiv preprint arXiv:1804.09541}.

\bibitem[Zhang et~al., 2019]{zhang2019dual}
Zhang, S., Zhao, H., Wu, Y., Zhang, Z., Zhou, X., and Zhou, X. (2019).
\newblock Dual co-matching network for multi-choice reading comprehension.
\newblock {\em arXiv preprint arXiv:1901.09381}.

\bibitem[Zhu et~al., 2018]{zhu2018hierarchical}
Zhu, H., Wei, F., Qin, B., and Liu, T. (2018).
\newblock Hierarchical attention flow for multiple-choice reading
  comprehension.
\newblock In {\em Thirty-Second AAAI Conference on Artificial Intelligence}.

\end{thebibliography}

\end{document}